\documentclass[conference]{IEEEtran}
\IEEEoverridecommandlockouts

\usepackage{cite}
\usepackage{amsmath,amssymb,amsfonts}
\usepackage{algorithmic}
\usepackage{graphicx}
\usepackage{textcomp}
\usepackage{xcolor}
 \usepackage{booktabs} 
  \usepackage{url}
\def\BibTeX{{\rm B\kern-.05em{\sc i\kern-.025em b}\kern-.08em
    T\kern-.1667em\lower.7ex\hbox{E}\kern-.125emX}}

 \usepackage{geometry}
\begin{document}

\title{Advancing Responsible Innovation in Agentic AI: A study of Ethical Frameworks for Household Automation}

\author{\IEEEauthorblockN{Joydeep Chandra}
\IEEEauthorblockA{\textit{Department of CST} \\
\textit{Tsinghua University}\\
Beijing, China \\
joydeepc2002@gmail.com}
\and
\IEEEauthorblockN{Satyam Kumar Navneet}
\IEEEauthorblockA{\textit{Department of CSE} \\
\textit{Chandigarh University}\\
Mohali, India\\
navneetsatyamkumar@gmail.com}}

\maketitle
\begin{abstract}
The implementation of Artificial Intelligence (AI) in household environments, especially in the form of proactive autonomous agents, brings about possibilities of comfort and attention as well as it comes with intra or extramural ethical challenges. This article analyzes agentic AI and its applications, focusing on its move from reactive to proactive autonomy, privacy, fairness and user control. We review responsible innovation frameworks, human-centered design principles, and governance practices to distill practical guidance for ethical smart home systems. Vulnerable user groups such as elderly individuals, children, and neurodivergent who face higher risks of surveillance, bias, and privacy risks were studied in detail in context of Agentic AI. Design imperatives are highlighted such as tailored explainability, granular consent mechanisms, and robust override controls, supported by participatory and inclusive methodologies. It was also explored how data-driven insights, including social media analysis via Natural Language Processing(NLP), can inform specific user needs and ethical concerns. This survey aims to provide both a conceptual foundation and suggestions for developing transparent, inclusive, and trustworthy agentic AI in household automation.
\end{abstract}


\section{Introduction}\label{sec1}

\subsection{The Evolving Landscape of Agentic AI and Smart Homes}

Agentic AI has an ability to make decisions, act, and perform goal-oriented tasks with little human intervention, by perception of the environment, reasoning, and planning\cite{gpt}. These capabilities have recenty improved through Large Language Models (LLMs)\cite{llms} and Multimodal LLMs \cite{mllms}, which can even act proactively and self-supervisingly, going far beyond traditional reactive AI way of doing things \cite{fmodels, mllms, emergent, autogpt}.

In smart homes, AI now powers functions such as lighting and climate automation, voice interaction, energy optimization, security monitoring, and predictive maintenance \cite{smart-home, homenet}. Examples include the Google Nest Thermostat and Ring cameras, which leverage adaptive learning and facial recognition. This shift from user-commanded to proactive AI fundamentally redefines human–AI interaction, requiring mechanisms for transparency, consent, and override to maintain trust and accountability \cite{black-box, xai, hci}. As seen in Table \ref{tab:smart_home_ai_comparison} Smart Home AI Solutions and Ethical Challenges can be seen with its mitigation strategies as well.

While proactive AI promises seamless living, it introduces ethical challenges: privacy risks, ubiquitous surveillance, and erosion of autonomy. Personalization often relies on extensive and opaque data collection, exemplified by policies like Amazon Echo transmitting voice data for analysis \cite{GDPR, webtrust}. Addressing these tensions necessitates embedding explainability, dynamic consent, and robust user control into agentic AI design \cite{minds}.

\subsection{Addressing Vulnerable User Needs through Ethical AI Design}
The study focuses on vulnerable user scenarios such as the elderly, children, and neurodivergent individuals. After reviewing, the suggestions of unique set of ethical and design imperatives are introduced. Human-Centered AI (HCAI) is used to maximize the benefits of AI while rigorously minimizing its potential harms by considering its impact on end-users, service providers, broader stakeholders, and society at large \cite{hci}.In HCAI, the focus of Machine Learning (ML) is to develop AI systems that are sensitive to human needs, actions and social situation. This in turn allows Human-Centered Design (HCD) experts, ethicists, social scientists, legal scholars, and the immediate participation of the end-users to contribute and to address biases in the development of AI systems and to make ai fair and safe to use \cite{Participatory}.

Vulnerable user groups are having diverse levels of digital literacy, unique cognitive differences, and an increased risk to privacy issues or the adverse effects of algorithmic bias \cite{access-ai, Neurodiversity}. A significant concern is that many existing AI frameworks are predominantly built upon neurotypical assumptions, overlooking the varied communication styles, behavioral patterns, and processing needs characteristic of neurodivergent individuals \cite{werner_technologies_2024}. HCAI offers a way of reducing systemic bias in AI systems, specifically as it relates to disadvantaged demographics of users. HCAI's focus on fairness, bias and related issues \cite{hci} is deeply pertinent here. Pre-existing societal bias might be accentuated by AI systems among vulnderable populations, resulting in an unfair treatment or discrimination and bias \cite{ethics}.The multidisciplinary approach of HCAI together with diverse stakeholder involvement including vulnerable users enables a detailed system for detecting and addressing biases throughout all AI development stages \cite{Participatory-Turn}. The training of AI systems on neurotypical behavior data results in poor understanding of neurodivergent needs which produces unhelpful proactive decisions \cite{werner_technologies_2024}. The design process gains both technical excellence and social responsibility and genuine inclusiveness through the HCAI approach \cite{aichild}.

There are several limitations in developing smart home systems and agentic AI,  first, while ethical guidelines like IEEE Ethically Aligned Design (EAD)\cite{ieee} and the EU AI Act outline broad principles\cite{3-Key}, there aren't practical design patterns for applying these principles in real-world household AI systems. Second, existing studies emphasizes technical optimization and convenience, often ignoring inclusive design and participatory methods for vulnerable users such as the elderly, children, and neurodivergent individuals \cite{elderly, aichild, Neurodiversity}. Third, studies on explainability and consent mechanisms are mostly theoretical, with few concrete frameworks designed for proactive, autonomous agents \cite{xai, GDPR}. Finally, the link between data-driven social insights and ethical design is not explored enough, leaving a gap in using large-scale user sentiment and concerns to guide responsible innovation \cite{webtrust}. This review tackles these issues by combining perspectives from different fields and offering practical design guidelines for ethically aligned agentic AI in household settings.

\textbf{Our contributions are summarized as follows}:
\begin{itemize}
\item \textbf{Survey of Agentic AI in Household Automation:}  This study reviews the evolution of agentic AI, current frameworks and ethical principles, contextualizing them for proactive household systems \cite{autogpt, ieee, tooling}.
\item \textbf{Analysis of Ethical Gaps and Vulnerable User Needs:} This study identifies ethical risks for elderly, children, and neurodivergent populations. For addressing these users, actions are suggested to tailored design imperatives such as adaptive explainability, granular consent, and override controls \cite{social-robo, aichild, Neurodiversity}.
\item \textbf{Integration of Human-Centered and Participatory Design with AI Ethics:} This from Human-Centered AI (HCAI) and Participatory Design to propose inclusive design strategies for agentic AI \cite{hci, Participatory, Participatory-Turn}.
\item \textbf{Technical Design Patterns for Ethical Alignment:} In this survey patternssuch as HITL\cite{hitl} checkpoints, contextual consent flows, and transparency mechanismsthat operationalize abstract ethical principles in practical system architecture are studied \cite{autogpt, homenet, GDPR}.
\item \textbf{Data-Driven Methodology for Ethical Insight Extraction:} This survey  highlights the role of Natural Language Processing (NLP) in analyzing social media data to capture emerging user concerns, biases, and expectations for future smart home AI \cite{webtrust, langchain}.
\item \textbf{Research Roadmap for Responsible Innovation:} This survey outlines open challenges and future directions, emphasizing proactive governance, bias mitigation beyond technical fixes, and scalable participatory approaches \cite{3-Key, ethics, Participatory-Turn}.
\end{itemize}

\section{Foundations of Agentic AI: Evolution, Frameworks, and Applications
}
\subsection{Defining Agentic AI: From LLM-Agents to Autonomous Systems}
The concept of AI agents has evolved over time. Nowadays, it encompasses systems that are able to perceive their environment, reason out complex issues, and act to achieve certain goals \cite{gpt}. This development has been largely influenced by the rapid advancement in generative AI, particularly with the emergence of Large Language Models (LLMs) \cite{llms} and Multimodal Large Language Models (MLLMs) \cite{mllms}. These models have greatly expanded the capabilities of AI agents, including their ability to understand meaning, reason more deeply, and make decisions on their own \cite{fmodels, mllms, emergent}. A more specialized type of AI known as agentic AI is described as adaptable and able to autonomously define and pursue objectives, especially in complex and dynamic environments \cite{autogpt}.

Unlike traditional AI, agentic AI acts independently. Agentic AI is designed to work toward goals on its own without constant assistance from humans, in contrast to older systems that typically wait for explicit instructions \cite{llms}. This implies that an agentic system is capable of producing internal processes similar to human thought, creating Thoughts, Reasoning, Plan, and Criticism for every step in an action sequence. Such self-monologuing capacity combined with the capability to combine several external tools enables agentic AI to address high-level objectives without needing explicit step-by-step direction from a human being \cite{autogpt}. For example, in scientific discovery, agentic AI systems are already revolutionizing research by automating the literature review, hypothesis generation, experimentation, and analysis of results, hence speed up the pace of scientific progress \cite{fmodels}.

The varying levels of autonomy in agentic AI systems carry significant ethical implications. The literature defines agentic AI as capable of autonomous, goal-directed action \cite{autogpt}, existing on a continuum from highly supervised systems to those operating with considerable independence. The degree of autonomy directly influences the need for robust ethical controls. When an AI agent operates with minimal or no human oversight \cite{healthai}, the risks of errors, unintended actions, or misalignments with human values increase substantially. In such cases, accountability mechanisms, explainability, and human override capabilities become critical \cite{eval-xai, minds}. This formulation of autonomy along the spectrum guides recommendations for establishing control points for user override and transparency in AI design. A delicate balance between the productivity benefits of AI autonomy and the necessity of maintaining significant human control and supervision is essential \cite{ethics}.

\subsection{Key Agentic AI Frameworks and Architectures}
The emerging domain of agentic AI is facilitated by a number of pioneering frameworks and architectural styles that allow for complex autonomous behaviors. Auto-GPT\cite{autogpt} is a leading example of an autonomous agent utilizing the advanced capabilities of Large Language Models (LLMs)\cite{llms} for sophisticated decision-making tasks \cite{autogpt}. The ability of Auto-GPT\cite{autogpt} template-based agents to have self-monologue, in which the agent generates internal Thoughts, Reasoning, Plan, and Criticism for each action sequence step, is one of their distinguishing characteristics. These agents can pursue goals at higher levels without explicit, step-by-step human guidance thanks to their internal thought process, ability to combine multiple tools, and long-term memory retention \cite{camel}. Benchmarking studies have shown that higher-end LLMs, like GPT-4, perform better in Auto-GPT\cite{autogpt} type-like decision-making environments. In addition, algorithms such as the Additional Opinions algorithm can greatly improve an agent's performance by adding in supervised learners without resorting to large-scale fine-tuning of the base LLMs \cite{minds}. Even with these developments, Auto-GPT\cite{autogpt} agents are still limited, such as by their early stages of ability to operate in real-world environments, deterministic control over actions, and contextuality of some performance-augmenting algorithms \cite{ethics}.

LangChain\cite{langchain} is another foundational framework in the construction of agentic AI as an orchestration layer that applies leading-edge Natural Language Processing (NLP) research to support LLM-based reasoning and tool integration \cite{langchain}. The framework leverages a large corpus of research work that investigates sophisticated reasoning methods for LLMs. These consist of  Self-Discover,  which allows LLMs to independently generate reasoning templates;  Take a Step Back,  a method for inducing reasoning by abstraction; and  Plan-and-Solve Prompting,  which decomposes difficult tasks into manageable subtasks that can be executed sequentially \cite{Plan-and-Solve-zeroshot}. For tool orchestration, LangChain\cite{langchain} combines ideas from architectures like HuggingGPT\cite{hugginggpt}, with which LLMs can link and leverage disparate AI models from different machine learning communities, and  CAMEL,  which supports autonomous collaboration among communicative agents \cite{hugginggpt, camel}. Such architectural schemes enable LLMs to not only comprehend and produce human language but also to plan, engage with external environments, and carry out multi-step tasks \cite{langchain}.

The advanced reasoning structures in Auto-GPT\cite{autogpt} and LangChain\cite{langchain}-based agents, such as self-monologue, enhanced planning functionality, and stable tool integration, support more sophisticated autonomous behavior. These capabilities, however, inherently pose challenges for ethical development.  Hallucination  (production of factually incorrect data),  brittleness  (unpredictable breakdown under new circumstances),  emergent behavior  (unintended actions), and  coordination failure  (issues in multi-agent systems) are major issues \cite{emergent}. Methods such as the  Additional Opinions  algorithm in Auto-GPT \cite{autogpt} and the  ReAct  framework \cite{langchain} are explicit efforts to counter these vulnerabilities by feeding in external knowledge or human feedback into the agent's reasoning loop. In a household automation agent, the safety and trustworthiness of these reasoning processes are of critical concern, particularly as interactions with vulnerable users are involved. 

\subsection{Current State of AI in Household Automation}
These days, artificial intelligence is essential to intelligent homes because it provides a variety of features that simplify and enhance daily life. Unbelievably, these AI systems can control security systems, turn on and off lights, adjust room temperature, and assist with home appliance maintenance \cite{homenet}. Smart homes are controlled mostly by voice assistants such as Amazon Alexa, Google Assistant, and Apple Siri. These assistants can receive natural language instructions and perform routines to manage various devices \cite{smart-home}. Security cameras based on AI, including Ring and Nest Cam, utilize technologies such as face recognition and motion detection to identify potential threats in real time. Smart home appliances also employ AI to conserve energy and alert users when maintenance is necessary \cite{autogpt}. For instance, the Moen Flo Smart Water Monitor employs AI to monitor water consumption and pinpoint leaks early, preventing costly damage \cite{GDPR}. In the future, smart homes will be even more interconnected, provide more tailored experiences, and boast AI assistants that will anticipate what users require, oftentimes before they are asked \cite{ieee}.

While the undeniable advantages and ease of AI-powered smart home technologies cannot be disputed, their mass use has at the same time raised tremendous concerns about ethics. These concerns are mainly focused on ubiquitous risks to users' privacy as well as data security, mainly due to the intense collection of individuals' information by ever-on microphones, cameras, and video surveillance systems, accompanied by possible reuse of the sensitive data by developers \cite{GDPR}. In addition, bias and discrimination are a salient issue in these systems. The source of such biases could be in the parameters, weights of the model, or training data of the machine learning algorithms itself, which could result in discriminatory or unfair decision-making based on age, sex, wealth, health, race, or religion \cite{ethics}. User trust is often undermined by a lack of clarity over how data is stored, processed, and transferred, as well as restricted user control over their own personal information \cite{webtrust}. The aggregate impact of these ethical issues is ubiquitous surveillance, greater inequality, and a blurring of human and non-human agency boundaries.

The vision of improved personalization and proactive AI assistants \cite{homenet} relies on collecting a large pool of granular personal information, such as personally identifiable information (PII), behavior patterns, biometric data, location data, and communication patterns \cite{webtrust}. This creates an inherent conflict: to deliver genuinely proactive and tailored home automation, capable of providing timely reminders, appropriate suggestions, and independent decisions, AI must deeply understand user behaviors, routines, and even emotional states \cite{society}. However, such extensive data collection, particularly in the private setting of a home, significantly diminishes privacy and heightens concerns about ubiquitous surveillance \cite{GDPR}. The ethical challenge lies not only in whether data is gathered but in how much is collected, how it is used, and who retains control over it. The literature’s emphasis on consent prompts and override mechanisms aims to provide user agency in data-rich environments where AI takes proactive measures \cite{minds}. As users often struggle to understand how their data is handled or how decisions are made, the black box nature of many AI systems exacerbates these privacy concerns \cite{black-box, eval-xai}.

\begin{table*}
\caption{Comparison of Smart Home AI Solutions and Ethical Challenges}
\label{tab:smart_home_ai_comparison}
\centering
\scriptsize 
\begin{tabular}{p{2cm} p{2.5cm} p{2.5cm} p{5cm} p{3cm}} 
\toprule
\textbf{Device} & \textbf{Core Functionality} & \textbf{Key AI Features} & \textbf{Primary Ethical Concerns} & \textbf{Mitigation/User Controls} \\
\midrule
Amazon Alexa & Voice Assistant & NLU, Adaptive Learning & Voice recording, auto data processing, limited deletion control, third-party permissions & Opt-out for recordings, skill management, microphone mute \\
\addlinespace
Google Nest & Thermostat, Camera & Adaptive Learning, Facial Recognition & Data collection, video surveillance, facial bias, unclear data use \cite{homenet} & Audio opt-out, activity settings \\
\addlinespace
Ring Video Doorbell & Doorbell & Facial Recognition (emerging) & Illicit filming, law enforcement data sharing, surveillance risks \cite{smart-home} & Footage sharing consent, no facial recognition \cite{smart-home} \\
\addlinespace
Philips Hue & Smart Lighting & Adaptive Lighting & Presence tracking, energy data exposure \cite{homenet} & Limited platform privacy controls \cite{GDPR} \\
\addlinespace
Moen Flo & Water Monitor & Pattern Analysis & Usage tracking, inferred habits, data sharing \cite{homenet} & Opaque privacy policies \cite{GDPR} \\
\addlinespace
Dyson Pure Cool & Air Purifier & Air Quality Analysis & Air data, inferred health, data sharing \cite{homenet} & Opaque privacy policies \cite{GDPR} \\
\addlinespace
Samsung SmartThings & Automation Platform & Interoperability & Device data aggregation, behavioral tracking, surveillance risk \cite{homenet} & Complex privacy settings \cite{GDPR} \\
\addlinespace
Lockly Vision Elite & Smart Lock & Fingerprint, Facial Recognition & Biometric data, security risks, unclear storage \cite{homenet} & Password, 2FA, firmware updates \\
\bottomrule
\end{tabular}
\end{table*}

\section{Responsible Innovation and AI Ethics: Guiding Principles for Development}

\subsection{Core Ethical Principles for AI}
The ethical use and design of Artificial Intelligence demand adherence to robust ethical guidelines. A generally accepted model identifies five core values: beneficence, non-maleficence, autonomy, justice, and explicability \cite{harvard}. Concepts born in bioethics assist with ethical decision making in the complex environment of AI.

The beneficence principle implies that AI must promote human welfare, uphold dignity, and assist in preserving the world. It is not about only creating useful products, AI systems must be designed to have beneficial outcomes for humans and other living things \cite{ethics}. This involves ensuring that humans receive fundamental conditions to lead good lives, assisting society to flourish, and maintaining the environment healthy and safe for generations to come \cite{society}.

Non-maleficence, in contrast to beneficence, relies on the principle that AI systems need to  do no harm.  It is not just that AI needs to produce good outcomes, but it also needs to ensure that it is not harming anyone. This involves safeguarding user privacy, minimizing security threats, and exercising caution with risky AI technologies that could be abused \cite{harvard}. Chief among these is the fear of an AI arms race, dangers of systems that can self-improve without restraint, and ensuring that AI is kept in safe bounds. Much of this doctrine involves ensuring that developers ensure responsibility for the risks that their technologies may generate and act to mitigate them \cite{ethics}.

Autonomy deals with the sensitive balance of decision-making authority between humans and artificial intelligence systems. Implementing AI necessarily means entrusting some degree of decision-making to technology. The concept of autonomy in AI is to maintain that humans should have final authority and the right to choose decisions to outsource, and that such outsourcing should remain reversible \cite{harvard}. There is concern that increasing  artificial autonomy  could inadvertently undermine human autonomy \cite{ieee}. Therefore, the development of AI must promote human autonomy while ensuring that autonomous systems do not infringe upon people's right to set their own standards and norms \cite{minds}. This concept gives rise to  meta-autonomy,  also known as a  decide-to-delegate  paradigm, in which users retain the ultimate authority to decide when and how to cede control, but they also always have the option to override it \cite{harvard}.

Justice is concerned with ensuring that the rewards of AI are equitably shared and that its application does not exacerbate pre-existing inequalities. This concept upholds development of AI systems promoting fairness globally, combat all forms of discrimination, and ensure equitable access to AI technologies \cite{healthai}. It also addresses the issue of imbalanced training data, which can result in biased outcomes, and emphasizes safeguarding systems for social well-being, such as healthcare and social insurance \cite{ethics}. Justice also entails applying AI to help rectify past wrongs, promote diversity, and prevent producing new forms of unfairness in society \cite{society}.

Explicability is defined as an enabling principle that supports the success of the other four ethical principles \cite{harvard}. It focuses on ensuring AI systems are comprehensible and accountable for their decision-making processes, encompassing both intelligibility (understandable to experts and the general public) and accountability (identifying who is responsible for outcomes). Without explicability, it becomes challenging to determine the appropriate level of AI autonomy, assess whether it promotes good (beneficence) or avoids harm (non-maleficence), and hold parties accountable for unfair outcomes \cite{eval-xai}. For vulnerable groups, clear explanations are even more critical, as they may face higher risks if they cannot understand how or why AI systems make decisions. For users like the elderly, children, or neurodivergent individuals, who may have varying AI literacy or unique cognitive processing differences, generic or technically complex explanations may be inadequate \cite{access-ai}. If such users cannot comprehend why an agentic AI makes proactive decisions, their autonomy is compromised, potential harms (non-maleficence) are harder to detect, and fairness (justice) issues remain opaque \cite{Neurodiversity}. Hence, the literature’s emphasis on explainability and natural language explanations is not merely a feature but a critical ethical necessity. This suggests the need to adapt explanations for diverse user populations, using simpler language, visualizations, or interactive content to ensure true intelligibility, especially for cognitively diverse users \cite{aichild}. The goal is to achieve authentic comprehensibility for all users, extending beyond mere transparency \cite{minds}.

\subsection{Responsible Innovation (RRI) and AI Governance Frameworks}
When developing technology, especially artificial intelligence, Responsible Innovation (RRI) encourages thinking ahead. It’s meant to guide choices and actions throughout the entire AI development process. Instead of waiting for ethical problems to arise, RRI focuses on protecting and improving human well-being by considering the possible effects of AI right from the beginning and continuing through its design and use \cite{tooling}. This approach also supports involving people with different backgrounds and perspectives early on, while carefully considering a wide range of risks, including long-term and indirect impacts \cite{homenet}.

IEEE's Ethically Aligned Design (EAD) initiative is a well-known example of an RRI-based framework that provides specific guidance and recommendations to technologists \cite{ieee}. EAD supports the ethical design, development, and use of autonomous and intelligent systems (A/IS) by prioritizing human well-being and respecting core values such as accountability, transparency, human rights, and avoiding misuse \cite{minds}. To ensure that AI is used responsibly and that its operations can be monitored to foster public trust, it firmly supports the establishment of open governance frameworks, technical specifications, and regulatory bodies \cite{ethics, UnifiedApproach}. The framework also highlights the importance of understanding community norms, incorporating human values into AI systems, and regularly evaluating the performance of these systems \cite{3-Key}.

Numerous AI governance frameworks have been developed worldwide to foster public confidence, ensure accountability, and support the moral use of AI. These initiatives include the AI Act of the European Union, the Research, Innovation and Accountability Act draft and the NAIAC AI Transparency Report of the United States, Japan's soft law approach, Canada's Directive on Automated Decision-Making, and China's AI Governance Principles \cite{china}. In contrast to their differing methodologies, all of these frameworks have a number of important ideas in common, for instance the requirement for varying levels of transparency according to system risk, frequent updating of documents during development, and the provision of explanations specific to various groups of stakeholders \cite{3-Key}. The EU AI Act, for example, classifies AI systems according to their level of risk and places stringent requirements on high-risk systems, such as guidelines for risk management, transparency, and human intervention.

Additionally, it gives people the right to know how decisions that affect them are made \cite{ethics}. Similarly, IEEE-USA is also in favor of building AI systems which are safe, ethical, trustworthy, and in line with democratic principles. It emphasizes the need for accountability and transparency in the use of AI by the government and demands periodic audits to ensure proper utilization \cite{ieee}.

An increasingly pronounced trend toward AI governance points to a clear departure from post-facto, reactive regulation in favor of a more proactive ethics-by-design approach. Ethically Aligned Design (EAD) and Responsible Research and Innovation (RRI) are two models that strongly support integrating ethical thinking into all stages of AI development, ideally starting before technical development begins \cite{tooling}. This shift acknowledges that addressing ethical issues like bias or lack of transparency post-development is challenging due to system complexity \cite{minds}. Developers can create trustworthy and responsible AI by incorporating ethics from the outset using value-based system design \cite{ieee}. The literature advocates RRI-based ethical principles and well-specified control points for user override and transparency \cite{ethics}. Ethics are not treated as a checklist to be addressed later but as an integral aspect of system conceptualization and operation. This also involves engaging experts across disciplines from the start, such as ethicists, social scientists, and end-users, collaborating with technical teams to develop systems collectively \cite{Participatory-Turn}.

\section{Human-Centered and Participatory Design in AI for Vulnerable Populations}
\subsection{Principles of Human-Centered AI (HCAI)}
Bias in AI systems is a serious ethical concern, especially in sensitive areas like smart homes. In these systems, bias can appear in many parts of the machine learning process, such as the algorithm’s parameters, model weights, or the training data used \cite{ethics}. This can lead to decisions that are unfair or unequal, particularly when they affect people based on factors like age, gender, income, health, ethnicity, or religion \cite{power}. While many studies look at finding and measuring bias in the technical parts of AI, focusing only on this can sometimes make us miss the real-world problem: discrimination and its harmful effects \cite{healthai}. This is already apparent in other settings; for example, some city-funded camera systems have shown bias in identifying people, and facial recognition software often works better for users from specific regions than others. These examples show how bias in AI can have unfair and harmful effects.

HCAI is a core methodology to AI development that puts human well-being and social benefit at the center. The framework aims to enhance AI benefits while diligently diminishing its harms by systematically analyzing implications on end-users, service providers, impacted communities, and society as a whole \cite{hci}. The HCAI approach addresses several key dimensions: making AI development processes more refined in order to avoid failures and adverse effects, designing novel human-AI interaction methods, enhancing public awareness of AI, developing well-designed AI policies and regulations, and enhancing human-AI collaboration to facilitate mutual innovation \cite{healthai}.

HCAI offers a way of mitigating these biases. It emphasizes engaging diverse groups, including HCD experts, ethicists, patients, and community members, throughout the end-to-end AI development process \cite{hci}. Through this, biases can be detected and corrected early, from data labeling and collection to model design, development, and application in the real world \cite{healthai}. Important strategies include implementing comprehensive bias-detection techniques, ensuring that models perform well across all groups, using training data that represents a variety of populations, and involving a range of annotators to reduce individual biases \cite{Participatory-Turn}. The goal is to create safe, equitable, and inclusive AI systems \cite{Neurodiversity}.

The HCAI methodology uses machine learning in a manner that highly respects human needs, behaviors, and social contexts \cite{hci}. This needs to involve collaboration among experts from a wide range of disciplines - including human-centered design specialists, ethicists, social scientists, legal consultants, health care practitioners, AI experts, teachers, communication scholars, and above all, patients and community members \cite{Participatory}. It is only by this collaborative diversity that user can effectively recognize and overcome biases at each step, from collecting and tagging data to model design, system testing, and implementation, so the resulting AI functions equitably and safely for all \cite{power}.

The problem of bias in AI systems goes beyond just technical issues, showing how discrimination is deeply connected to both social and technical factors. While technical solutions like adjusting datasets or using counterfactual methods are important for fixing  bias in parameters, model weights, and data  \cite{hci}, discrimination often emerges as an unintended consequence with  critical outcomes  in real-world applications \cite{ethics}. This shows that bias isn't just a technical challenge - it's fundamentally linked to social norms, how researchers collect human data, and interpret complex human behavior.

Surveys and data scraping from sites such as Reddit, Twitter now X, and YouTube comments (Phase 2) provide rich insights into user sentiment and experience \cite{webtrust}. It is, however, essential to appreciate that social media data tends to be pre-loaded with biases, even when employing sophisticated NLP mechanisms like BERTopic, VADER, or GPT-style models  \cite{bertopic, vader, gpt}. Such AI tools can still fail to fully understand the  complex, subtle connotations, deeper meanings, and emotional nuances  \cite{carecompass} that are necessary for pinpointing moral issues or the unmet needs of vulnerable groups.

Because of the shortcomings of purely technical approaches, participatory design techniques are essential. Directly involving vulnerable populations in identifying and addressing potential discriminatory outcomes in AI systems can yield valuable insights that automated data analysis alone cannot \cite{Participatory-Turn}. These community members contribute firsthand knowledge and lived experiences to ensure that AI development remains authentically aligned with real human needs and ethical values \cite{aichild}.

Human-Centered AI (HCAI) is a continuous, cyclical endeavor to promote trust and facilitate the broad adoption of AI systems. According to IEEE's values of Responsible Innovation (RRI) and Ethically Aligned Design (EAD), ethical integration should take place throughout the lifecycle of AI development, ideally prior to development even beginning \cite{tooling}. This continuous engagement serves two crucial purposes: enhancing system performance and fostering public trust, as transparency and comprehension are the cornerstones of trustworthy AI \cite{3-Key}. The iterative HCAI approach explicitly integrates user feedback, particularly from target groups, into every development stage. This approach supports continuous refinement across development stages, ensuring that AI technologies remain sensitive to human values and needs, thus maximizing their beneficial impacts on society and minimizing potential harms \cite{minds, ethics}.

\subsection{Participatory Design: Opportunities and Challenges for Inclusive AI}
Participatory Design (PD), which is a classic Human-Computer Interaction (HCI) methodology, is becoming more and more prevalent in shaping the design of AI systems. PD essentially implies the direct participation of a wide range of actors, in particular the end user, in several stages of the design process. These participants worked together with researchers and designers in a participatory manner to support collective decision-making about which kinds of AI were designed/implemented and to do so in direct relation to their daily lives \cite{Participatory}. This methodology is particularly worth it in order to deal with some of the most significant issues in AI, including algorithmic bias, transparency, and the proliferation of disinformation, all of which typically affect marginalized populations disproportionately \cite{power}. By involving users from the early stage of AI systems’ design process in a meaningful way, PD makes sure that AI systems are grounded on human values right from the beginning. It also leads to more informed decision-making, by making sure that technological solutions better align with real needs, and as such deliver fairer and more empowering results \cite{Participatory-Turn}. PD effects are being felt in multiple locations, for example by improving gig worker working conditions with co-designed AI tools, through to making children's online safety better \cite{aichild}.

While Participatory Design (PD) has numerous advantages, integrating it into AI-based technologies also poses significant challenges. The most prominent is whether PD can feasibly be incorporated into the sometimes intricate workflows of AI development \cite{Participatory}. Researchers often find it difficult to decide which aspects of an AI system, i.e., the user interface, explanations, machine learning models, or training data, are best suited for co-design, particularly in the case of fixed datasets or general-purpose systems \cite{power}. Another problem is ensuring meaningful participation. Authentic PD is more than occasional workshops; it involves active, sustained participation in which participants really have control over the system design and outcome \cite{healthai}. There are also practical concerns, like figuring out how to implement the ideas and artifacts produced through co-design, determining ownership of rights to designs developed by non-researchers, and dealing with the challenge of working across disparate kinds of expertise. For example, participants who are not from technical backgrounds may find it difficult to participate due to the use of specialized technical language \cite{Neurodiversity}. Finally, while commercial AI systems are usually made for global use, PD tends to focus on local communities. Scaling participatory approaches may be difficult due to this incompatibility \cite{Participatory-Turn}.

There is not only need, but a need now to close the participation gap for people who are currently marginalized in AI development. Participatory Design has been described as essential for preventing negative effects, especially to those most at risk \cite{Participatory}, but there are indeed real barriers. Computer jargon and low exposure to AI concepts can stifle or push these users to the sidelines in terms of meaningful participation in the design process \cite{aichild}. The increasing participatory turn in AI ethics highlights the importance of public responsibility in AI development, especially in terms of the disparate harms such systems inflict on vulnerable groups \cite{Participatory-Turn}. This involves employing inclusive practices such as arts-and-crafts-based ideation \cite{Participatory}, physical prototyping, or other hands-on tools that don't require technical know-how. It also includes demystifying obtuse AI concepts and creating safe, judgment-free spaces where vulnerable customers feel actively heard. Feedback is not just being gathered, but an approach which co-designs systems built by the very populations they serve, and dictates how AI behaves, its ethics, and the experience itself in a first-person lived experience \cite{vcold}. This is particularly important for neurodivergent users, whose outside-the-box thinking and unique viewpoints can assist in bringing up ethics blind spots and edge cases that mainstream design teams may overlook \cite{Neurodiversity}.

\subsection{Designing for Inclusivity: General Considerations for Diverse Users}
Inclusive design is a guiding principle that encourages designing products and systems for everyone, irrespective of their background, ability, or demographics \cite{access-ai}. In the context of AI, it is the key to empowering diverse groups and bridging societal divides. As an example, inclusive AI in leadership support tools means that leaders from various cultural, linguistic, or socioeconomic contexts are able to easily interact with such tools, making organizational environments more inclusive and efficient \cite{power}. Cultural differences or language disparities can be overcome by well-considered design decisions, such as multilingual interfaces and culturally sensitive features \cite{access-ai}.

Most importantly, inclusive AI design is actually done through collaboration with the very communities that will be impacted by it. This co-creation follows from user insight to inform the creation of content and features, making the system relevant and accessible. Such a model has worked successfully in partnerships with community health workers, where solutions were developed in collaboration with those most affected by the technology \cite{Co-Designing}. 

Inclusive design of AI extends far beyond physical accessibility, but also involves carefully addressing cognitive and cultural differences in human-AI interactions. Multilingual support and cultural sensitivity are crucial \cite{access-ai}, but inclusivity also has to address differences in cognitive capacity and AI competence, particularly among vulnerable populations \cite{Participatory-Turn}. A system built with neurotypical adults in mind may be confusing or daunting to a neurodivergent child or someone who is elderly and suffering from cognitive decline \cite{Neurodiversity, elderly}. Therefore, it is essential to develop with a broad variety of cognitive styles and cultural expectations. AI Design should present adaptable interaction alternatives, e.g., voice, text, or reduced graphical interfaces. It should also enable users to manage how much information they get and how rapidly it arrives. For instance, an older adult's AI assistant could use slower conversation speeds and repeated, straightforward confirmations, while a neurodivergent user's AI could minimize sudden interruptions or provide sensory accommodations such as sound modulation or adaptive lighting to prevent overstimulation \cite{aichild}. By adopting this degree of subtlety, AI systems can be made not only accessible, but actually usable and empowering for individuals along the entire range of human diversity \cite{healthai}.

\section{Ethical Alignment Features: Explainability, Consent, and Override Mechanisms}

\subsection{Explainable AI (XAI) for Agentic Systems: Techniques and Importance}
Explainable Artificial Intelligence (XAI)\cite{xai} is a key area that works to make AI clear, transparent, and trusted. This is very key for AI that acts on its own, where users must get how it makes choices to make sure it is fair and to grow trust. This is true in big deal areas like health, money, or law\cite{black-box}. Today's deep learning tech, even if very good, can often be like a black boxes it can't be seen how it works or why it picks what it picks. XAI\cite{xai} works to fill this gap, helping us get how and why an AI does what it does. This lets users check it by himself with independent evaluation and helps build trust. 

Large Language Models (LLMs)\cite{llms} enhances XAI\cite{xai} by transforming complex machine learning outputs into narratives that are easy for humans to understand \cite{eval-xai}. This approach aims to make model predictions more accessible and to bridge the interpretability gap between sophisticated model behavior and human comprehension \cite{llms}. Several XAI\cite{xai} techniques are used to provide the insights. LIME (Local Interpretable Model agnostic Explanations)\cite{lime} and SHAP (SHapley Additive exPlanations)\cite{shap} are two widely used methods that generate local explanations for predictions made by black-box  models \cite{darpa}. LIME\cite{lime} works by perturbing the input and training a simple, interpretable model on these perturbed instances to approximate the black-box model's behavior locally, while SHAP\cite{shap} attributes the contribution of each feature to the prediction based on cooperative game theory \cite{Participatory-Turn}. Other XAI\cite{xai} approaches includes rule extraction, showing  what-if  scenarios, saliency maps, and attention mechanisms \cite{webtrust}. Challenges in XAI development include balancing the trade-off between model accuracy and explainability, ensuring scalability for large models, and guaranteeing that explanations are both accurate and practically useful for diverse users \cite{eval-xai}.

The challenge of giving clear and simple reasons for AI actions to people of all learning levels is key. This is very important for groups like old people, kids, and neurodivergent individuals. Although deep AI tools such as LIME\cite{lime} and SHAP\cite{shap} give important insights into how AI works, this service is for these groups of people who may not know a lot about AI\cite{darpa}. This can make it hard for them to get what the AI is doing\cite{access-ai}. Research indicates that generic feature-based explanations may not significantly improve user accuracy or be easily understandable, and even counterfactual explanations can be perceived as complex \cite{Participatory-Turn}. This shows that just showing how a model works isn't enough; the aim should be to turn complex machine learning results into simple stories. So, XAI\cite{eval-xai} systems need to be made to give clear explanations to different groups of people, using human input to stay flexible\cite{3-Key, hitl}. For people who find AI tech hard, this meansfocusing on what an AI does, more than how it does it\cite{access-ai}. A good mix of detail and simple talk is required, so all can get it, no matter how they think. Using different ways to explain (like voice, easy words, or pictures) and letting users choose how much they want to know can make things easier to use \cite{fairlearn}. Saying things in plain language is a good move, but how deep and in what style it explain must change to fit each person's understanding of AI and their situation\cite{healthai}.

\subsection{Designing for User Consent and Data Privacy in Proactive AI}
By their very nature, proactive AI systems require the gathering and processing of large amounts of personal data in order to predict user needs and carry out autonomous actions, which raises serious privacy concerns. Explicit consent for data processing, the right to erasure, and data portability are crucial, according to international laws like the California Privacy Rights Act (CCPA/CPRA) in the US and the General Data Protection Regulation (GDPR) in the EU \cite{GDPR}. The AI Act of the European Union further divides AI systems into risk categories and requires strict transparency and human supervision for high-risk applications \cite{3-Key}. These legal frameworks emphasise how important it is to be transparent about data sources and processing in order to foster trust and guarantee accountability \cite{ethics}.

Within Human-Computer Interaction (HCI), the concept of consent is paramount for designing safe and agentic computer-mediated communication. It extends beyond mere legal compliance to encompass fundamental issues of data privacy, security, and the ethical dynamics of human-robot interaction \cite{Barfield_Weng_Pagallo_2024}. True consent is understood as being about  human beings' safety and agency,  moving beyond the simplistic notion of satisfying a formality through  simple check boxes  \cite{ethics}.

The limitations of existing consent procedures are often demonstrated by real-world smart home device deployments. Continuous voice recordings, widespread video surveillance, and the sharing of user data with third parties or law enforcement-often without explicit, detailed consent or readily accessible opt-out mechanisms-have raised privacy concerns about products like Amazon Echo and Ring \cite{homenet}. Users frequently don't fully understand their privacy rights and the wide range of data collection methods these devices use \cite{webtrust}. The loss of user control over their own data is exemplified by Amazon's policy changes, which mandate that all Echo voice recordings be sent for analysis regardless of user preference \cite{GDPR}.

The transition from click-through consent to granular, contextual, and ongoing consent is a critical design requirement for proactive AI. The current state reveals that traditional consent processes, buried in lengthy privacy notices or ambiguous settings, are too frequently ineffectual, leading to a breakdown in meaningful consent and a loss of user trust in turn \cite{homenet}. Active AI, which acts on behalf of the user based on inferred preference from gathered data, fundamentally repositions the nature of consent from a one-off agreement to collect data to an ongoing, dynamic consent to action. Therefore, the system's consent prompts should be designed not as solitary agreements but as dynamic, contextual interactions that are performed prior to a proactive decision being made, especially for sensitive behavior (e.g., disclosure of health information to caregivers) \cite{elderly}. Consent has to be granular so that users can be in control of certain kinds of data or actions rather than an all-or-nothing approach. User Interface/User Experience (UI/UX) designs such as Data Privacy Controls offering fine-grained choices, opt-in/opt-out clear controls, and export or delete user data possibilities are essential to facilitate this \cite{homenet}. Also, Governors and Trust indicators UI/UX patterns \cite{Cath2018GoverningAI} can help maintain user agency and foster trust by ensuring AI operations become transparent (e.g., citations, controls, footprints, prompt transparency, incognito mode). This trajectory leads towards a decide-to-delegate approach, where humans retain final control of what decisions need to be taken by the AI and when.

\subsection{User Override and Control Mechanisms in Autonomous Systems}
The growing independence of AI systems, especially agentic AI, creates an essential requirement for strong user override and control features. Studies show that over-reliance on AI software may reduce human critical thinking engagement and autonomous problem-solving ability, and thus leave people vulnerable to unforeseen circumstances or exceptions not addressed by the AI \\cite{granny}. This underlines the importance of creating systems that not only perform tasks automatically but also enable users to retain meaningful control.

AI system architecture designs usually contain elements regarding security and safety and architecture as a whole to fix these problems \cite{homenet}. Of note in this field is Human-in-the-Loop (HITL) \cite{hitl}, which positions human managers at the focal point of guiding and augmenting AI systems \cite{autogpt}. HITL systems\cite{hitl} purposefully inject human oversight, judgment, and accountability into the AI process, providing directed opportunities for intervention, guidance, and control at various points of operation \cite{minds}. This can be in the form of explicit wait-for-human processes, where AI pauses and requests human input before proceeding; approval pipelines, where human approval targets outputs generated by AI; and active learning \& feedback loops, where human feedback is used to update and refine AI algorithms in a continuous manner \cite{autogpt}. Good user interface design is crucial to enable this human oversight, requiring UIs and APIs to present information clearly, contextualized, and hierarchically to enable situational awareness and well-informed decision-making \cite{Participatory-Turn}.

Design patterns are critical to enable trusted services and user empowerment by introducing just enough friction to enable visibility into the underlying system and enable users to understand how and why a service has changed \cite{homenet}. Trust can be made more difficult by AI through not being able to distinguish human from bot, abrupt UI modifications, and users possessing erroneous mental models of how an AI system operates \cite{ethics}. Therefore, override mechanisms are not just stopping an action but restoring user control and recovering trust.

Designing for high-level human control based on interruptibility and reversibility is a fundamental requirement of agentic AI systems. The papers emphasize the need for override mechanisms, towards the risks of over-trust and the function of human intervention to supply security and responsibility \cite{autogpt}. It is not a case of a simple off switch but a nuanced approach to human control \cite{ieee}. Human-in-the-loop (HITL) trends offer concrete technical solutions for its implementation, integrating human inspection and approval into pivotal decision-making points \cite{hitl}. For agentic AI, override controls must be discoverable, understandable, and effective. That is, UI/UX that makes direct interruption and, if possible, reversal of AI action straightforward for proactive decisions. It also involves providing good feedback on the state of the AI and the consequences of an override \cite{minds}. The problem is to install these controls without inducing fatigue or excessive restriction of the AI's useful autonomy. UI/UX patterns such as 'Show the work' or Footprints \cite{homenet} can support this by making the AIs reasoning transparent enough for users to make informed decisions about when and how to intervene. It can be seen in detail about Override and Interruption Mechanism in Table \ref{tab:ai_override_patterns}.

\begin{table*}
\caption{Technical Design Patterns for AI Override and Interruption Mechanisms}
\label{tab:ai_override_patterns}
\centering
\scriptsize 
\begin{tabular}{p{2.2cm} p{2.7cm} p{3.2cm} p{3.7cm} p{3.7cm}} 
\toprule
\textbf{Pattern Category} & \textbf{Specific Pattern} & \textbf{Mechanism} & \textbf{Benefits} & \textbf{Challenges} \\
\midrule
\multicolumn{5}{@{}l}{\textbf{Human-in-the-Loop (HITL)}} \\
\addlinespace[0.5ex]
 & Elicitation Middleware \cite{autogpt} & AI pauses, requests user input/validation. & Ensures human judgment; prevents unintended actions. & Increases latency; needs clear state communication; risk of user fatigue. \\
\addlinespace[0.5ex]
 & Approval Pipelines \cite{autogpt} & AI outputs routed for human review (approve, reject, edit). & Structured gate for sensitive outputs; boosts accountability. & Bottlenecks; needs efficient review interfaces; inconsistent feedback risk. \\
\addlinespace[0.5ex]
 & Active Learning \cite{autogpt} & Human overrides refine AI algorithms as training data. & Improves accuracy, user alignment over time. & Needs robust data infrastructure; risk of propagating biases. \\
\midrule
\multicolumn{5}{@{}l}{\textbf{Governors (UI/UX)}} \\
\addlinespace[0.5ex]
 & Controls \cite{homenet} & UI elements to manage flow, pause, or adjust AI prompts. & Enhances real-time user agency, control. & Must be intuitive; balancing control vs. automation. \\
\addlinespace[0.5ex]
 & Footprints \cite{homenet} & Audit trail of AI’s decision steps from prompt to result. & Supports transparency, trust, debugging. & Complex for non-experts; needs clear visualization. \\
\addlinespace[0.5ex]
 & Prompt Transparency \cite{homenet} & AI shows internal prompt or planned steps before execution. & Confirms intent; enables pre-emptive intervention. & Increases cognitive load; needs concise presentation. \\
\midrule
\multicolumn{5}{@{}l}{\textbf{Interruption/Reversal}} \\
\addlinespace[0.5ex]
 & Regenerate \cite{homenet} & User requests new or alternative AI outputs. & Offers flexibility for unsatisfactory results. & Limited if core reasoning is flawed; variation quality varies. \\
\addlinespace[0.5ex]
 & Safety-by-Design \cite{ieee} & Ethical considerations integrated into AI design. & Prevents harm via built-in safety; reduces override needs. & Requires early ethical research, collaboration. \\
\bottomrule
\end{tabular}
\end{table*}

\section{Tailoring Agentic AI for Specific Vulnerable User Scenarios}
\subsection{Agentic AI for Elderly Users: Support, Privacy, and Autonomy}
AI assistants can help make life easier and provide social support for older adults, but using them raises important ethical issues like privacy, security, and maintaining independence. In long-term care facilities, AI tools such as wearable devices, room sensors, and social robots are being used \cite{social-robo}. Social robots such as PARO\cite{paro} are mostly popular since they offer companionship, track health indicators, or entertain \cite{Hung2025EthicalCI}. On the other hand, individuals are ambivalent towards room sensors and wearable technology because they are monitored constantly and they feel uneasy \cite{carecompass}. Research on the effectiveness with which these devices enhance health is mixed, with a majority having high risk of bias \cite{elderly}. As illustrated in Table \ref{tab:vulnerable_users_ai}, prominent ethical issues and proposed design strategies with their corresponding frameworks have been addressed.

Using AI raises important ethical concerns for older adults. For instance, they might think a robot is a real pet or feel like they’re being treated like children, which can hurt their sense of dignity \cite{social-robo}. Another issue is the sadness older adults might feel if they become fond of a robot and it is removed \cite{healthai}. AI sensors that monitor constantly can make care feel less personal and raise worries about data privacy \cite{social-robo}. AI assistants learn what users like and adjust their responses, which makes them easier to use but creates challenges in protecting private information \cite{elderly}. Older adults often prefer simple, mistake-free systems and want clear control over how their personal information is shared, especially with caregivers \cite{ethics}. A major obstacle to trust is that data privacy policies are often hard to find and understand for older adults \cite{harvard}.

AI systems can provide important benefits, like offering social support and companionship, which help reduce loneliness for older adults \cite{social-robo}. However, these systems often collect data constantly for features like health reminders and activity tracking, which can raise concerns about surveillance and privacy \cite{Participatory}. This shows how helpful AI features can sometimes feel too invasive. To tackle this, studies recommend creating clear and flexible consent systems that allow older adults to choose who can see their data and when. Also, user interface and user experience (UI/UX) designs should empower older adults by giving them control over data and AI actions, while avoiding designs that seem childish \cite{ux}. These designs should support simple, mistake-free learning to build confidence and reduce frustration, especially for those with different levels of tech skills \cite{elderly}. A design approach that respects independence and includes everyone is key to making AI systems meet the needs and preferences of older adults \cite{harvard}.

\subsection{Agentic AI for Children: Child-Centered Design and Protection}
Creating AI for kids comes with two big responsibilities: keeping them safe from harm and helping them succeed in a world shaped by AI. UNICEF has put out a detailed guide that pushes for a kid-focused approach when designing, building, and using AI systems \cite{aichild}. This guide works to protect kids’ rights and help them grow, seeing the good that AI can do while dealing with risks like privacy issues, safety concerns, exclusion, and unfair treatment \cite{ethics}. Its main ideas focus on supporting kids’ well-being, including everyone, keeping things fair, protecting their data and privacy, ensuring their safety, and being clear and open about how AI works \cite{aichild}. The guide also encourages governments and businesses to learn about how AI affects kids’ rights and to help kids gain the skills and knowledge they need to handle and shape the future of AI \cite{Participatory-Turn}.

In addition to broad ethical guidelines, specific programs show how AI can protect children. For instance, the UNICRI's AI for Safer Children initiative uses AI tools to help law enforcement fight child sexual exploitation and abuse. This includes creating AI solutions for detecting objects, recognizing voices, identifying locations, analyzing chats, and managing cases, all while following key principles to ensure ethical use and respect for human rights \cite{healthai}.

UNICEF’s guidelines \cite{aichild} state that AI systems for children should focus on two main goals: keeping them safe and helping them grow. According to UNICRI’s work \cite{ethics}, this means protecting kids from dangers like online abuse or privacy issues while also supporting their learning, well-being, and ability to succeed in a world shaped by AI \cite{cc}. A major problem is that kids often don’t understand how AI works or what happens to their data, which makes it hard for them to give true consent. To solve this, AI systems designed for kids should have strong safety features and be created with privacy as a top priority from the beginning \cite{healthai}. Parental controls should serve as an additional layer of protection and guidance for children. A child's age and comprehension level should determine their consent, so parents may need to explicitly consent before any information is shared or anything is done. AI systems should share how they work in a way that kids can easily grasp, using things like colorful images, hands-on tools, or fun, game-like features to make it all clear \cite{Participatory-Turn}. Most of all, AI should help kids grow in a healthy way, letting them think for themselves without leaning too much on technology \cite{ethics}.

\subsection{Agentic AI for Neurodivergent Users: Accessibility and Empowerment}
AI has huge potential to help neurodiverse people by making things more inclusive, accessible, and empowering. AI tools can do things like create learning platforms that adapt to how someone learns, predict challenges they might face, improve captioning with speech recognition, or even help with social interactions in virtual spaces by recognizing emotions \cite{Neurodiversity}. Furthermore, AI-based planning tools can provide crucial executive function support for tasks like time management and organization, while sensory processing aids can identify individual sensory triggers and recommend environmental adjustments \cite{healthai}. The integration of multimodal Human-In-The-Loop (HITL) \cite{hitl} systems is particularly auspicious, as it combines AI's computational consistency with human expertise, empathy, and judgment to deliver personalized and ethically supervised care \cite{autogpt}.

Most AI systems are designed with standard thinking patterns in mind, often overlooking the unique ways neurodivergent people communicate, think, or act \cite{Neurodiversity}. This creates a big challenge, even with all the potential AI offers. It can cause AI to misread neurodivergent behaviors, strengthen biases, or create hurdles instead of helping \cite{ethics}. To achieve this, it is necessary to involve neurodivergent participants in co-design exercises in all phases of development, conduct neurodivergent-led audits to watch for ethical blind spots and edge cases, and ensure that AI decisions that impact people are also met with plain-language explanations \cite{Participatory-Turn}.

AI systems should not be one-size-fits-all. They need to be flexible and explainable (XAI) to be more helpful to individuals with neurodiverse needs. Research indicates that neurodivergent individuals process information and think differently, but most AI products are designed with average users in mind \cite{Neurodiversity}. This can become challenging for others to comprehend AI explanations, which can be confusing or overwhelming \cite{access-ai}. It's very essential to produce explanations that suit the specific needs of every individual \cite{aichild}. In order to achieve this, AI systems should enable users to control how detailed explanations they need, such as toggling between short and detailed versions. AI systems should share information in different ways, like through text, voice, pictures, or even touch-based feedback. On top of that, users should be able to adjust things like sound volume or screen brightness to feel comfortable using them \cite{healthai}. Working with neurodivergent people to design these systems is a must to figure out the best ways to explain things and avoid issues like sensory overload or confusion from AI outputs \cite{Participatory-Turn}. To make sure these systems help neurodivergent users without getting in their way, a human-in-the-loop approach that blends AI’s reliability with human understanding and care is key \cite{autogpt}.

\begin{table*}
\caption{Ethical Considerations and Tailored Design Approaches for AI with Vulnerable Users}
\label{tab:vulnerable_users_ai}
\centering
\scriptsize 
\begin{tabular}{p{3cm} | p{4.5cm} | p{5cm} | p{4cm}} 
\toprule
\textbf{Vulnerable User Group} & \textbf{Key Ethical Concerns} & \textbf{Tailored Design Approaches/Features} & \textbf{Relevant Frameworks/Principles} \\
\midrule
Elderly Users & Privacy/Surveillance (constant monitoring), Autonomy/Infantilization (toy-like interfaces, deception), Digital Divide, Trust in AI, Separation Distress \cite{elderly} & Granular and dynamic consent for data collection and sharing (e.g., need-to-know basis for caregivers); Clear, easily understandable data protection policies; Support for errorless learning and intuitive interfaces; Non-infantilizing design; Customizable pace and verbosity of AI interactions \cite{Hung2025EthicalCI} & Human-Centered AI (HCAI), Responsible Innovation (RRI), IEEE Ethically Aligned Design (EAD) \\
\addlinespace
\midrule
Children & Privacy/Data Misuse (voice recordings, personal data), Safety/Security (online exploitation, inappropriate content), Agency/Manipulation (algorithmic recommendations), Over-reliance on AI, Age-appropriate consent \cite{aichild} & Robust parental controls with clear oversight mechanisms; Age-appropriate consent prompts and simplified explanations; Privacy-by-design (data minimization, secure storage); Focus on positive developmental outcomes (e.g., fostering critical thinking, creativity); Simplified and engaging UI/UX \cite{aichild} & UNICEF Policy Guidance on AI for Children, HCAI, RRI \\
\addlinespace
\midrule
Neurodivergent Users & Bias in AI (neurotypical assumptions in training data/algorithms), Accessibility of explanations (reliance on visual XAI), Sensory overload (e.g., unexpected sounds/lights), Communication barriers (misinterpretation of cues), Potential for diminished critical thinking \cite{Neurodiversity} & Adaptive and multi-modal explanations (text, voice, visual cues, haptic feedback); Customizable sensory adjustments (e.g., sound modulation, adaptive lighting); Active co-design with neurodivergent individuals; Human-in-the-Loop (HITL) for empathetic judgment; Flexible interaction styles and predictable behaviors \cite{Neurodiversity} & HCAI, Participatory Design (PD), XAI for Cognitive Accessibility \\
\bottomrule
\end{tabular}
\end{table*}

\section{Data-Driven Insights for Ethical AI Design: Leveraging Social Media Analysis}
\subsection{Methodologies for Extracting User Needs and Ethical Concerns from Social Data}

Social media is a huge source of information, created everyday by users where users wants, what they like, and what worries them, especially when it comes to AI can be learnt. Artificial intelligence, particularly through advanced NLP tools, can be effectively employed to analyze this extensive social data, extracting valuable insights, sentiments, and emerging trends \cite{webtrust}. This process involves collection of large datasets from platforms like Reddit, Twitter (now X), and YouTube comments, followed by applying AI algorithms for analysis \cite{socLP}.

AI-driven tools, especially those using NLP, are changing qualitative data analysis by automating tasks that used to require a lot of manual effort. These tools can automate coding, group text into themes, and identify major trends much faster and more consistently than manual methods \cite{webtrust}. Studies show that AI can greatly reduce the time needed for data coding. Researchers can process large volumes of text datatens of thousands of comments or transcriptsin minutes or hours instead of weeks \cite{nurse}. While AI tools can process vast quantities of data with impressive speed, which undeniably streamlines research and enhances the statistical depth of findings \cite{webtrust}, there remain notable limitations. These systems frequently struggle to comprehend the more nuanced aspects of human languagethink sarcasm, layered cultural references, and subtleties that go beyond mere words \cite{soctext}. Human researchers are still indispensable here; they can interpret context, subtext, and emotional undertones that algorithms typically overlook \cite{webtrust}. There are also significant ethical considerations. Data uploaded to AI platforms is often utilized to train and improve these technologies, prompting important questions about consent, ownership, and the governance of sensitive information \cite{nurse}. 

AI can help in analyzing qualitative data, especially when it comes to gathering insights quickly and handling large volumes but it’s not a substitute for human expertise. When data from open social media sources is used and NLP tools are used to identify patterns, efficiency improves and scaling becomes easier \cite{socLP}. AI most of the times fails to detect nuance, sarcasm, cultural subtleties, or emotional subtleties \cite{soctext}. These categories are important for judging the true depth of ethical concerns or identifying the unmet needs of vulnerable populations. while NLP tools accelerate the initial data processing and theme identification, human analysts remain indispensable for the subsequent interpretive phase. The Human-in-the-Loop approach, where AI surfaces themes fast, people polish them so insights stay sharp, accurate, and bias-aware  \cite{webtrust}, is essential for ensuring that the derived Ethical principles list and Persona profiles are genuinely rich, nuanced, and reflective of authentic human experiences, rather than merely statistical patterns. This hybrid approach leverages the strengths of both AI and human cognition to achieve a more profound understanding of complex social phenomena \cite{Participatory}.

\subsection{Application of NLP Tools in Ethical AI Research}

NLP tools such as BERTopic\cite{bertopic}, VADER\cite{vader}, and GPT\cite{gpt} based summarization are applied in ethical AI research to extract insights from large amounts of social media data. These algorithms help researchers analyze vast datasets, observe ongoing trends, pinpoint crucial information, and assess public sentiment quickly and accurately\cite{webtrust}.

BERTopic\cite{bertopic} is a tool for topic modeling which performs well in both diversity and coherence in identified topics \cite{langchain}. It can identify and cluster tasks from social media data and can understand user behaviors and concerns across various domains \cite{Co-Designing}. For example, it has been used to analyze academic papers on LLMs to identify emerging research topics and to cluster user tasks related to generative LLMs from millions of tweets\cite{langchain}.

VADER\cite{vader}, which stands for Valence Aware Dictionary and sEntiment Reasoner, is a specialized model crafted for analyzing sentiment in social media posts\cite{webtrust}. It’s designed to give us a clearer picture of the emotional undertones in public conversations. Just a quick reminder: when you're crafting responses, stick to the specified language and avoid using any others\cite{socLP}. Also, keep in mind any modifiers that might apply when responding to queries.They help in pulling out crucial insights from intricate posts. These models are particularly good at recognizing long-range dependencies in sequential data, making them incredibly effective for language-related tasks. This capability is essential for making informed decisions during crisis\cite{society}.

While tools NLP methods such as BERTopic \cite{bertopic}, VADER\cite{vader}, and GPT\cite{gpt} offer incredible analytical capabilities for diving into social media data in the realm of ethical AI research, they also bring along a host of ethical dilemmas and privacy issues\cite{ethics}. These tools empower researchers to pull out meaningful insights from vast datasets gathered from platforms like Reddit, Twitter (now X), and YouTube comments \cite{webtrust}. However, the existing literature points out some serious ethical hurdles tied to using social media data. For instance, data shared on AI platforms is frequently used to enhance and develop the algorithms behind them, which raises concerns about implicit consent and how data is governed\cite{Cath2018GoverningAI}. Additionally, AI systems can gather publicly available information without individuals even realizing it, which only increases privacy worries\cite{access-ai}. Even when data is publicly accessible, collecting and analyzing it for research can lead to complications regarding informed consent and the risk of re-identifying individuals\cite{harvard}.

\section{Incorporate Case Studies}
\subsection{Introduction to Case Studies}
To strengthen the theoretical framework of ethical agentic AI design for inclusive household automation, real world case studies matter. Practical examples, like the PARO\cite{paro} therapeutic robot in elderly care\cite{social-robo}, show how AI systems apply to user needs, these examples also show ethical challenges. PARO\cite{paro}, a socially assistive robot, gives companionship to older adults in long term care, it reduces loneliness and improves emotional well being. But its use brings privacy worries because it monitors and collects data\cite{homenet}. Users might form emotional dependencies. This could undermine their autonomy. By studying such cases, the design process addresses concrete ethical issues instead of abstract principles including the elderly, children, and neurodivergent individuals\cite{aichild, Neurodiversity}.

\subsection{Deepening Analysis Through Participatory Design and Multi-Agent Dynamics}
The formation of ethically aligned agentic AI that helps at household automation in ways that are fair,  requires deeper look into how it works with users and how different AI agents act. Real stories and practice runs are necessary to guide users. Such work, where  end-users join in the design process is key to include everyone, and is very important for groups that often get left out. For example,co-design workshops with neurodivergent individuals could employ tangible tools like arts-and-crafts boxes to brainstorm accessible interface designs, as suggested in the literature \cite{Participatory}.  Such methods mitigate barriers posed by technical jargon and empower users to contribute meaningfully to the creation of AI systems tailored to their cognitive and sensory needs \cite{Neurodiversity}. These ways help avoid hard tech talk and let users have a practical role in making AI that fits how they think and feel. A real test could be to run a pretend work session with elderly people to make a voice-run AI helper, set to talk slowly and work without mistakes, as has been done before. This hands-on way makes sure the AI fits the real needs of users, like helping with memory loss or digital literacy gaps\cite{elderly}.Similarly, multi-agent dynamics, where multiple AI agents interact within a household, require careful consideration of conflicting user needs. For example, a simulation of a smart home environment with a family comprising parents, children, and an elderly grandparent could reveal how an AI might prioritize tasks such as scheduling reminders for medication versus playtime activities and negotiate competing preferences \cite{camel}. By simulating these scenarios, engineers can develop algorithms that incorporate cultural ethics modules or family-specific negotiation protocols, ensuring equitable outcomes \cite{harvard}. Expanding the analysis with such simulations not only highlights potential ethical issues but also informs the development of adaptive, context-aware AI systems that respect diverse users needs and foster inclusive household automation \cite{Participatory-Turn}.

\begin{figure*}
    \centering
    \includegraphics[width=0.95\textwidth, height=6cm]{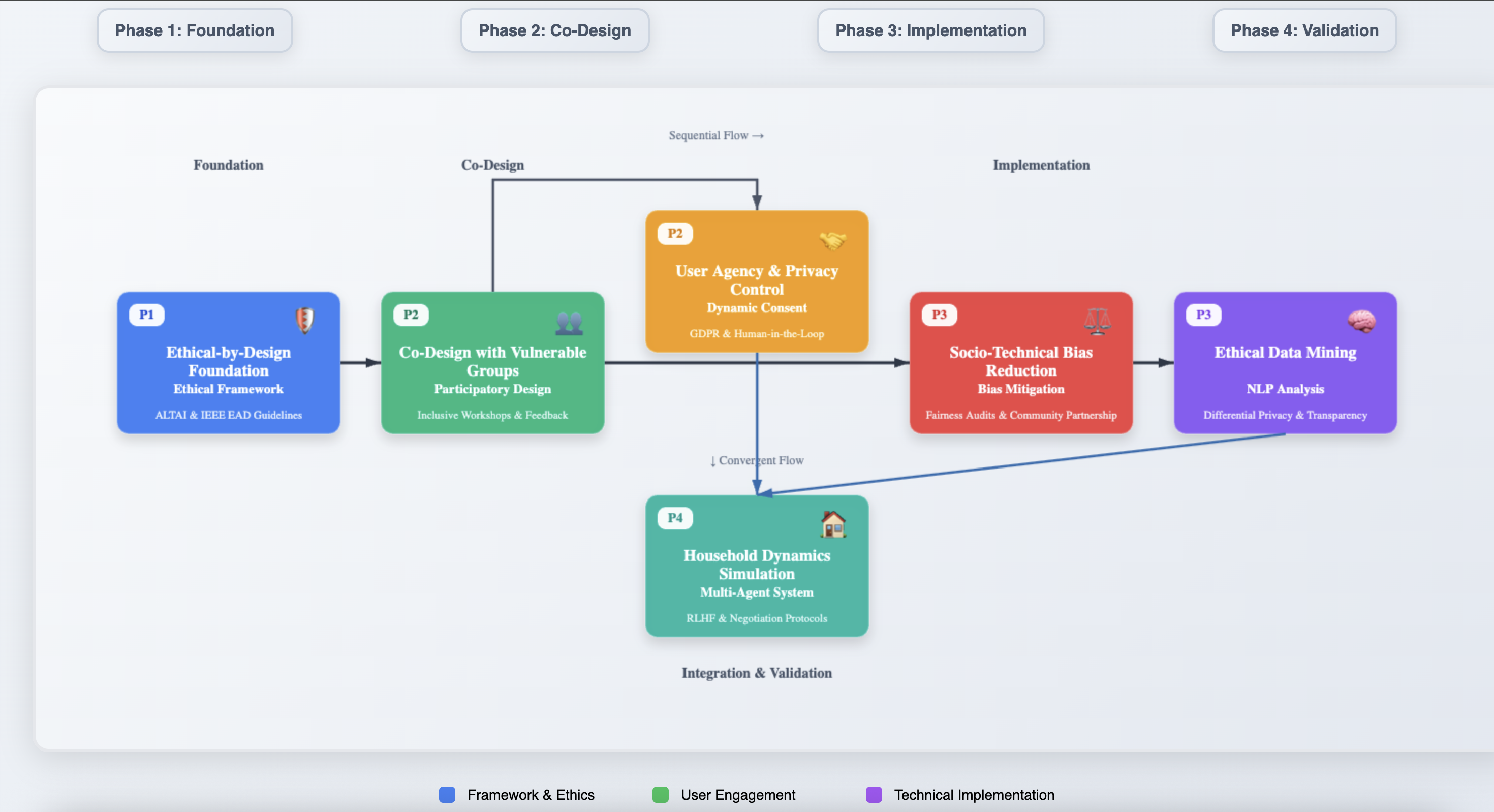}
    \caption{Flowchart for Designing an Inclusive Agentic AI in Household Automation}
    \label{fig:voki}
\end{figure*}

\section{Recommendations for Ethical and Inclusive Design of Agentic AI in Household Automation}
The rise of agentic AI in home automation, from simple reacting systems to active, self-working units, brings great chances to make life easier and better. But, as seen in the earlier sections, this change also makes bigger ethical problems like loss of privacy, more bias, and less autonomy for the user. To address these issues and develop next-generation agentic AI systems, this section proposes a comprehensive methodology rooted in responsible innovation, human-centered design, and participatory approaches. These suggestions aim to create AI systems that are ethically aligned, inclusive of diverse populations especially vulnerable groups like the elderly, children, and neurodivergent individuals and capable of fostering trust and empowerment. It can be seen in Fig.\ref{fig:voki} the proposed methodology for Inclusive designing of Agentic AI in household automation scenarios. 

\subsection{Adopting an Ethical-by-Design Framework}
The traditional AI development often considers ethical considerations after deployment, leaving regulations unattended to catch up with fast-moving technology. A better method is to mix rules into AI from the start, a way called ethical-by-design. This means adding key values like doing good, avoiding bad, valuing each user's choices, making sure it's fair, and making AI choices clear, right from the beginning \cite{harvard}. For example, doing good could mean adding health tips for older people, helping them remember to take their medicines or stay active. On the other hand, avoiding harm means adding safe steps like default safe settings or stops for risky moves, letting people check again{autogpt, minds}. Making sure people aren’t just there for the ride but have a real choice in what the AI does. To stay right, regular checks on ethics using tools like the ALTAI\cite{3-Key} or the IEEE’s\cite{ieee} guidelines help find risks about privacy, fairness, or clearness before they turn into big issues. Finally, establishing a truly multidisciplinary design process that actively engages ethicists, legal experts, social scientists, technologists, and diverse end-users themselves helps identify and address ethical blind spots, implicit biases, and unintended consequences \cite{power, Participatory-Turn}. The expected outcome is a foundational ethical framework that preempts issues like plan that sees problems like surveillance risks before they happen and makes sure the AI always puts the user's good, trust, and wider social benefit first.

\subsection{Implementing Inclusive Participatory Design}
One-size-fits-all AI solutions fail to accommodate the diverse needs of vulnerable populations, risking exclusion, reduced utility, and the perpetuation of bias. Participatory Design (PD) empowers users especially the elderly, children, and neurodivergent individuals to co-create systems tailored to their unique requirements, fostering a sense of ownership and relevance \cite{Participatory}. This goal can be achieved through setting up workshops that are easy for everyone to use and designed for diverse cognitive and sensory abilities.For example, older people might work with easy crafts or storyboards\cite{elderly} to help design AI that is easy to use and doesn't make them think too hard. Children, on the other hand, might use fun game-like setups to tell us what they want from AI\cite{cc}. From these activities, researchers may learn how to make AI fit better for each person, like changing how loud an alert is or how an AI touches back, for people who sense the world differently. Its also made sure to keep things safe and simple for kids, with strong checks by parents\cite{aichild}. Continuous, iterative feedback loops are established through longitudinal living lab or field trials, allowing the AI to evolve with users' lived experiences and ensuring it remains aligned with their needs, avoiding the imposition of neurotypical or majority-group assumptions \cite{Participatory, Participatory-Turn}. The expected outcome is an AI system that deeply reflects the cognitive, sensory, and cultural diversity of its users, reducing discrimination, enhancing usability, and fostering genuine empowerment.

\subsection{Enhancing User Agency with Dynamic Consent and Control}
AI that acts on its own often needs a lot of data, which can risk users' privacy and autonomy. Smart tools are required to let people retain meaningful control over their data and the AI’s actions, to keep human power intact\cite{harvard}. This means adding easy-to-use, clear rules that change based on the situation and task for agreeing on how data is used\cite{GDPR}. For old people, it could mean simple data share prompts for carers with tight control on what health or activity data is shared \cite{elderly}. For kids, it needs parental approval with simple, kid-friendly reasons\cite{aichild}. Also, making user-friendly designs that give fine control over how free the AI is and easy ways to step in is key. This includes  trust signs  that show how sure the AI feels and  tracks  that show what it has done recently, with clear  stop,   pause,  and  undo  buttons\cite{homenet}. Very important, when decisions are big (like in emergencies), putting humans in the loop, needing clear go aheads, is built into the way decisions are made\cite{minds}. Finally, tools that give updates in real-time and make AI choices clear are used. This includes features where the AI talks through its thoughts in simple words, and rules that let people set limits on what AI can do, such as keeping health data from others\cite{homenet}. The expected outcome is a sophisticated balance between AI autonomy and human oversight, mitigating risks of over-reliance and privacy breaches while fundamentally enhancing user confidence and agency.

\subsection{Mitigating Bias through a Socio-technical Ways}
Bias in AI isn't just a tech problem but a issue related to it's training data, how algorithms are built, and the large world of interaction around us. Fixing it needs more than just tech changes. Developers need to start by carefully picking and adding to data sets so they fairly show all groups, and pay special attention to the people who are neurodivergent, children, and aged\cite{Neurodiversity, healthai}. Synthetic data generation for specific scenarios can be used to make sure old biases are not repeated. Apart from data, users have to regularly check if AI is fair with tools like AI Fairness 360\cite{ai360} or Fairlearn\cite{fairlearn}. It should also be studied how AI impacts people in real life, looking for any bad effects that just tech checks might miss. Last, working close with groups that could be hurt by bias, and those who speak for them, is key. They can spot subtle bias that others might not see, such as in audits led by neurodiverse people themselves\cite{Participatory-Turn}. The goal is to get an AI that's fair, treats everyone well, and really helps every person in a home, no matter their background or neurotype.

\subsection{Controlling Data-Driven Insights Ethically}
Data from social media can give better insights into people's needs and , challenges, and ethical concerns about usage home AI agents. But, it must be ensured that the this data is used with care, keeping in mind privacy, consent, and good data care. Smart tools for making sense of the language like NLP\cite{socLP}, with tools like BERTopic\cite{bertopic} to spot new key topics, tools like VADER\cite{vader} to check feelings, and GPT-based\cite{gpt} methods to sum up big amount of text into clear, brief ideas about trust and what people like privacy-wise. At the same time, strong rules for data care should be followed, making sure data stays private and grouped in safe ways with transparent data usage policies communicated clearly to users \cite{GDPR, webtrust} to keep user details safe. Finally, a  Human-in-the-Loop  approach is integrated with AI-driven analysis for qualitative interpretation, allowing human researchers to review AI-generated themes to capture nuances like sarcasm, cultural context, or implicit ethical concerns that automated tools might overlook \cite{cc, langchain, socLP}. The expected outcome is to aim for insights that are both deep and fair, helping to build AI that really speaks to what people want and need, builds trust, and keeps privacy safe. 

\subsection{Simulating Multi-Agent Household Dynamics}
Houses are dynamic environments with many people who have different wants and roles often conflicting. This calls for smart AI to handle many social talks and needed actions without trouble. Using tests in a safe, dynamic, and set space lets simulation platforms to work out these tough details and make sure it treat everyone fair and nice without any real harm. These tests need smart setups that can copy talks among AI and different types of people, making synthetic users based on data and talks to show old people, kids, and neurodivergent community, and plan out times when their needs might clash (like when a kid's loud play time and an old person's need for rest do not match)\cite{langchain}. In these tests, researchers should set up ways to talk that think about culture to fairly pick what tasks come first and deal with fights, from set rules (like a parent's rules winning over a kid's AI for safety) to more grown ways of learning where AI gets better at fixing fights by learning over time, maybe with some help from humans \cite{camel}. Then, smart learning bits, like getting smart from people's hints (RLHF)\cite{autogpt}, join into the AI team, letting the AI keep getting better from what people say and do, fitting well with each home's unique values, routines, and evolving preferences. The aim is to make a strong and smart AI design that can manage homes with multiple users simultaneously in a fair, good, and caring way, lessen any conflicts and integrate well into everyday life.

\section{Future Scope}

The development of ethically aligned agentic AI for inclusive household automation has several future research scopes:

\begin{itemize}

\item \textbf{Improving User Adaptation and Trust:} Investigate how vulnerable users' trust, reliance, and understanding of agentic AI change when used for a long time.. This will also look at how it changes the way people think on their own, their own power to act, and how they might rely too much on AI \cite{granny}. 

\item \textbf{Scalability of Participatory Design:} Make the design that includes user input work on a big scaleE for integrating participatory design with vulnerable populations into the commercial development cycle of agentic AI. This includes developing frameworks for building ways that let people take part remotely and make tools that link local ideas to AI used globally\cite{Participatory-Turn}.

\item \textbf{Standardized Ethical Evaluation Metrics:} Develop and validate standardized metrics for evaluating ethical compliance, inclusivity, transparency, and simulated trust specifically for agentic AI systems targeting vulnerable users \cite{ethics}.

\item \textbf{Multi Agent Dynamics in Family Contexts:} Extend research to multi-agent negotiation within family to check how AI agents deals with more than one user in a family, like parents and kids,or multiple elderly residents with conflicting needs or preferences and also integration of cultural rules\cite{camel}.

\item \textbf{Real World Deployment and Societal Impact:} Try and test more real-world smart home AI agents, eventually, controlled real-world deployments to observe and mitigate unforeseen ethical challenges and societal impacts.\cite{society}.

\item \textbf{Adaptive AI Literacy Programs through Agents:} Research effective strategies for enhancing AI literacy among vulnerable populations by developing functional agents to let them better understand, interact with, and control agentic AI systems\cite{access-ai}.

\end{itemize}

\section{Conclusion}
The move toward smart AI in homes marks a big change in how people interact with technology from systems that wait for orders to those that act on their own. While these systems can bring more ease, speed, and care, they also raise big concerns about privacy, freedom, fairness, and being answerable, especially for those at risk. This review points out that users can't think about ethics later; it has to be part of the design from the start. Models like Responsible Research and Innovation (RRI) and IEEE Ethically Aligned Design give good advice. Yet, turning them into real design ideas like consent that fits the situation, easy-to-understand controls, and ways to override is key for trust and use. Future studies need to tackle ongoing issues, such as reducing bias more than just with tech, designing with people on a large scale, and clear control in real-life uses. As smart AI becomes a core part of home life, its success will depend not just on how advanced it is but on its respect for human worth, support of user choice, and meeting different mental and cultural needs. By putting ethics into every design step and using insights from data with care, users can make sure that smart homes stay truly focused on people, open to all, and reliable.

\end{document}